\documentclass[runningheads,a4paper]{llncs}

\usepackage{amssymb}
\usepackage{graphicx}

\usepackage{url}
\urldef{\mailsa}\path|mohamed.daoudi@telecom-lille.fr|
\urldef{\mailsb}\path|{stefano.berretti, pietro.pala, alberto.delbimbo}@unifi.it|
\urldef{\mailsc}\path|yvonne.delevoye@univ-lille3.fr|
\newcommand{\keywords}[1]{\par\addvspace\baselineskip
\noindent\keywordname\enspace\ignorespaces#1}

\begin{document}

\mainmatter  

\title{Emotion Recognition by Body Movement Representation on the Manifold of Symmetric Positive Definite Matrices}

\titlerunning{Emotion Recognition by Body Movement Representation}

%
%
\author{Mohamed Daoudi$^1$
\and Stefano Berretti$^2$\and Pietro Pala$^2$\and Yvonne Delevoye$^3$\and Alberto Del Bimbo$^2$}
\authorrunning{M. Daoudi et al.}

\institute{$^1$IMT Lille Douai, Univ. Lille, CNRS, UMR 9189 -- CRIStAL -- Centre de
Recherche en Informatique Signal et Automatique de Lille, F-59000 Lille, France\\
$^2$University of Florence, Florence, Italy\\
$^3$Univ. of Lille, CNRS, UMR 9193 SCALab, France\\
}

%
%

\toctitle{Emotion Recognition by Body Movement Representation on the Manifold of Symmetric Positive Definite Matrices}
\tocauthor{M. Daoudi et al.}
\maketitle

\begin{abstract}
Emotion recognition is attracting great interest for its potential application in a multitude of real-life situations. Much of the Computer Vision research in this field has focused on relating emotions to facial expressions, with investigations rarely including more than upper body. In this work, we propose a new scenario, for which emotional states are related to 3D dynamics of the whole body motion. To address the complexity of human body movement, we used covariance descriptors of the sequence of the 3D skeleton joints, and represented them in the non-linear Riemannian manifold of Symmetric Positive Definite matrices. In doing so, we exploited geodesic distances and geometric means on the manifold to perform emotion classification. Using sequences of spontaneous walking under the five primary emotional states, we report a method that succeeded in classifying the different emotions, with comparable performance to those observed in a human-based force-choice classification task. 
\end{abstract}

\section{Introduction}
Automatic analysis of human motion has been an active research topic for several years, with outcomes that have been beneficial to a number of different applications, including security surveillance, health-care at home, athletes training and natural interfaces, to say a few.
The variety in human body (size, height, corpulence), in the way different people perform an action, and even in the way a same person performs one action at different times, makes the task of human motion analysis very challenging. 
In the last decades, a consolidated line of research has analyzed the human motion from RGB and depth data enabling tasks such as action and gesture recognition~\cite{herath:2017}. 
However, body movements carry a multitude of information, also indicative of our intentions, 
inter-personal attitudes, expectations and emotions. 
Of particular interest are basic emotions (i.e., \textit{anger}, \textit{disgust}, \textit{fear}, \textit{happiness}, \textit{sadness}, and \textit{surprise}) that are innate in all humans and are cross-culturally recognizable. 
These basic emotions can be further clustered in \emph{active} (anger, happiness, surprise) and \emph{passive} (fear, sadness, disgust).
Recently, the study of computational models for human emotion recognition has gained increasing attention not only for commercial applications (to get feedback on the effect of advertising material), but also for gaming and monitoring of the emotional state of operators that act in risky contexts such as aviation. Most of these studies have focused on the analysis of facial expressions, but important clues can be derived by the analysis of the dynamics of body parts as well.

The first rigorous investigation on the expression of emotions through the body dates back to Darwin's seminal work on ``The expression of the emotions in man and animals''. 
Since then, research in the field of Emotional Body Language (EBL) has addressed this subject from both a bio-mechanical 
and a psychological perspective. 
The recognition of emotions from the analysis of body movements entails a higher level of complexity; indeed, since the body is primarily used to perform manipulative actions and enable motion, emotional clues can only be detected as secondary signatures on top of those ongoing actions. 
Hence, most EBL studies have addressed only the question of what aspects of 3D body kinematics are impacted by emotional states. Such studies have reported that body rhythmicity is slower for low energy emotions (\textit{sadness} and \textit{fright}) and faster for high-energy emotions (\textit{anger}); these patterns have been confirmed across a variety of natural actions, e.g., door knocking, 
walking, 
and dancing. 
Nevertheless, such behavioral findings are not sufficient to tackle the difficult question of emotion classification through body motion observation.

Finding a compact and effective representation of body movement is a difficult task when considering the complexity of temporal dynamics. In addition, measuring the similarity between two temporal sequences for the purpose of classification is complicated in itself. In fact, the Euclidean distance is unsuitable for comparing temporal sequences, and Dynamic Time Warping is often used as an alternative~\cite{muller:2007}. To address these issues, there is a recent trend that investigates matrix based solutions. The idea of these methods is to embed the non-linearity of the sequence into a matrix representation, then exploit the geometric properties of the space (manifold) the matrices lay in to perform distance measurement and classification.
Examples are the block Hankel matrix~\cite{bhattacharya:2014}, and the Gram matrix~\cite{zhang:2016}. 
%
Along this line of research, \textit{covariance matrices} have found success in several computer vision applications, including activity recognition, visual surveillance and diffusion tensor imaging. 
Recently, several properties of the covariance matrices have been popularized by investigating the related Riemannian manifold of Symmetric Positive Definite matrices (SPD)~\cite{jayasumana:2015}.


Based on the above considerations, in this paper we propose a new solution to perform human emotion recognition from the analysis of the temporal dynamics of the joints of the body skeleton in the 3D space. Human motion is captured by the evolution across time of the 3D position of the joints 
in an appropriate reference system. Then, a covariance matrix descriptor is extracted from the features across the sequence frames. Exploiting the properties of the covariance matrix, this descriptor is mapped to the non-linear Riemannian manifold of SPD matrices. Finally, emotion classification is performed on the manifold by computing geodesic distances between test sequences and template emotions obtained as the average on the manifold of training examples. Experiments show the potential of the proposed solution, which obtains comparable results to those scored by human evaluators. In summary, the contributions of this work are:
\textit{(i)} Analysis of the dynamics of the full-body movement to understand human emotions 
over long sequences, while most of existing works use body-parts and short time; 
\textit{(ii)} A representation of the body movement that uses the covariance descriptor to capture the dynamics of the skeleton joints, and analyzes these descriptors in the related Riemannian manifold of SPD matrices. 
This is obtained by the adoption of a suitable distance measure and mean computation to perform classification on the manifold.

The rest of the paper is organized as follows: Previous work related to the proposed method is summarized in Sect.~\ref{sect:related-work}; In Sect.~\ref{sect:spd}, we present the mathematical background for the non-linear Riemannian manifold of SPD matrices; In Sect.~\ref{sect:representation}, the adopted representation of the joints of the skeleton and its movement is presented; The classification approach on the manifold is discussed in Sect.~\ref{sect:classification}; Results and a comparative evaluation are reported in Sect.~\ref{sect:results}; Finally, conclusions and future work directions are drawn in Sect.~\ref{sect:conclusions}.

\section{Related Work}\label{sect:related-work}
%
The decreasing cost of whole-body sensing technology and its increasing reliability, make it possible to investigate 
the role played by body expressions as a powerful affective communication channel. 
Kapur et al.~\cite{kapur:2005} were among the first to address these aspects in 3D. Using a Vicon Motion Capture system, they collected gestural sequence data depicting sadness, joy, anger, and fear emotions of five subjects. The 3D position of 14 markers, plus their velocity and acceleration were calculated, and the mean values of velocity and acceleration and the standard deviation values of position, velocity and acceleration across the sequence were considered as descriptors. Finally, classification was performed comparing five different classifiers. 
Gong et al.~\cite{gong:2010}, addressed the problem of recognizing affect from non-stylized human body motion using 3D joints of the skeleton. Motion capture data were represented by a descriptor based on the shape of signal probability density function, and SVM were used for classification. Experiments were performed on a dataset of 30 individuals performing knocking, throwing, lifting and walking motions in four affective states (i.e., neutral, happy, angry and sad).
%
Karg et al.~\cite{karg:2010} analyzed the human gait to reveal persons affective state, comparing inter-individual versus person dependent recognition. The dynamics of the body was captured by measuring features such as the stride length, cadence, velocity, minimum mean and maximum values of angles between body parts. Then, these features were reduced using PCA, kernel PCA, LDA and GDA techniques, while classification was performed with NN, Naive-Bayes and SVM.
Results showed that recognition is highly affected by individual walking styles and individual expressions of affect (accuracy of 69\% and 95\% were reported for the inter-individual and person dependent case, respectively, based on the observation of a single stride). 
They also observed that automatic recognition based on gait patterns tends to 
better recognize \textit{active} than \textit{passive} emotional states. 
%
For a comprehensive coverage of the topic, we refer to the survey by Kleinsmith and Bianchi-Berthouze~\cite{kleinsmith:2013} that reviewed the literature on affective body expression perception and recognition,
and the survey by Karg et al.~\cite{karg:2013} that summarized methods to recognize affective expressions from body movements, and the converse problem of generating movements for virtual agents or robots, which convey affective expressions. 

%
%

Several works used the special Riemannian manifold 
of SPD matrices. One typical case for which such matrices arise in practice is when covariance descriptors are used to model image sets or temporal frame sequences in videos. 
Covariance features were first introduced by Tuzel et al.~\cite{tuzel:2006} for texture matching and classification. 
Several studies have extended the use of covariance descriptors to the temporal dimension, with application to human action and gesture recognition.
%
Sanin et al.~\cite{sanin:2013}, proposed an action and gesture recognition method from videos based on spatio-temporal covariance descriptors. Prior to classification, points on the manifold were mapped to an Euclidean space, through Riemannian Locality Preserving Projection~\cite{harandi:2012}. 
%
%
Bhattacharya et al.~\cite{bhattacharya:2016} constructed covariance matrices, which capture joint statistics of both low-level motion and appearance features extracted from a video. 
To facilitate the classification task, matrices were mapped to an equivalent vector space obtained by the matrix logarithm operation, 
which approximates the tangent space of the original SPSD space of covariance matrices. Then, human action recognition was formulated as a sparse linear approximation problem, in which these mapped features are used to construct an overcomplete dictionary of the covariance based descriptors built from labeled training samples.
%
%
In~\cite{faraki:2016}, Faraki et al. noted that when covariance descriptors are used to represent image sets, the result is often rank-deficient. Most of the existing methods solve this problem by accepting small perturbations to avoid null eigenvalues and thus, employ standard inference tools. What they proposed, instead, were novel similarity measures specifically designed for 
the particular case where symmetric matrices are not full-rank (i.e., Symmetric Positive Semi-Definite matrices, SPSD).
%

\section{Manifold of Symmetric Positive Definite Matrices}\label{sect:spd}
Let $f$ ($f \in \mathbb{R}^d$) be a $d$-dimensional feature vector of landmarks, and $D_{d \times n }= \lbrack f_1, \cdots, f_n \rbrack$ denote a set containing the $d$-dimensional feature descriptors of $n$ images of an image set. The covariance matrix $\textbf{C}$ of the set is defined by:
\begin{equation}
\displaystyle{\textbf{C}=\frac{1}{(n-1)}\sum_{i=1}^n (f_i- \mu)(f_i-\mu)^T} \; ,
\end{equation}

\noindent where $\mu$ is the sample mean. A non-singular covariance matrix of size $d \times d$ belongs to the set of symmetric positive-definite (SPD) matrices. These do not form a vector space (the space is not closed under matrix subtraction), rather they form a connected Riemannian manifold $Sym^{+}_d$~\cite{Bhatia2007}. As such, the distance between SPD matrices is not accurately captured by the Euclidean distance. Covariance matrix has recently received increasing attention in Computer Vision by leveraging Riemannian geometry of SPD matrices. 

Indeed, several distance measures on $Sym^{+}_d$ have been proposed.
The most widely used is the Log-Euclidean Riemannian Metric (LERM)~\cite{arsigny:2007}. Given two covariance matrices $\textbf{C}_1$ and $\textbf{C}_2$, their LERM is computed as:
\begin{equation}
\label{eq:LERM}
d(\textrm{\textbf{C}}_1,\textrm{\textbf{C}}_2)=\| \log(\textrm{\textbf{C}}_1)-\log(\textrm{\textbf{C}}_2) \|_{\textrm{F}} \; ,
\end{equation}

\noindent where $\|\cdot\|_{\textrm{F}}$ is the Frobenius norm, and $\log(\textrm{\textbf{C}})$ is the matrix logarithm of $\textbf{C}$.

\section{Representation of Body Movement}\label{sect:representation}
The dynamics of body movements is expressed by a sequence of observation vectors capturing the position of body joints across time. More specifically, the human body is approximated by a skeleton composed of $N_J$ joints. Accordingly, the posture of the body at a generic observation time $t$ is expressed by a vector $p\in \mathbb{R}^{3N_J}$ composed of the $(X,Y,Z)$ coordinates of body joints at time $t$:
\begin{equation}
p(t) = \left[ x_1, y_1, z_1,\ldots,
x_{N_J}, y_{N_J}, z_{N_J} \right] \; .
\end{equation}

In order to also keep track of the body dynamics at each observation time, the posture vector is augmented with the velocity vector that is composed of the $(X,Y,Z)$ components of the velocity of body joints at time $t$:
\begin{equation}
v(t) = \left[ v_{x_1}, v_{y_1}, v_{z_1},\ldots,
v_{x_{N_J}}, v_{y_{N_J}}, v_{z_{N_J}} \right] \; .
\end{equation}

The velocity of a generic joint at time $t$ is computed by finite difference of joint positions at time $t$ and $t-1$, assuming zero velocity at $t=0$.

In order to make the position and velocity vectors invariant to the orientation of the body with respect to the camera, coordinate values $(X,Y,Z)$ are normalized by expressing them in a skeleton centered coordinate system $(X_S,Y_S,Z_S)$. This is computed as the orthonormal basis resulting from the PCA of the positions of the torso joints at $t=0$. 
A compact yet representative description of the dynamics of body movements across a temporal observation window $[0,T]$ is extracted by computing the covariance matrix of the concatenated posture/velocity vectors. This results into a symmetric $6N_J \times 6N_J$ square matrix.

\begin{figure}[t!]
\centering
\includegraphics[width=0.4\linewidth, viewport=3.4cm 0.7cm 18.4cm 28.6cm, angle=-90, clip]{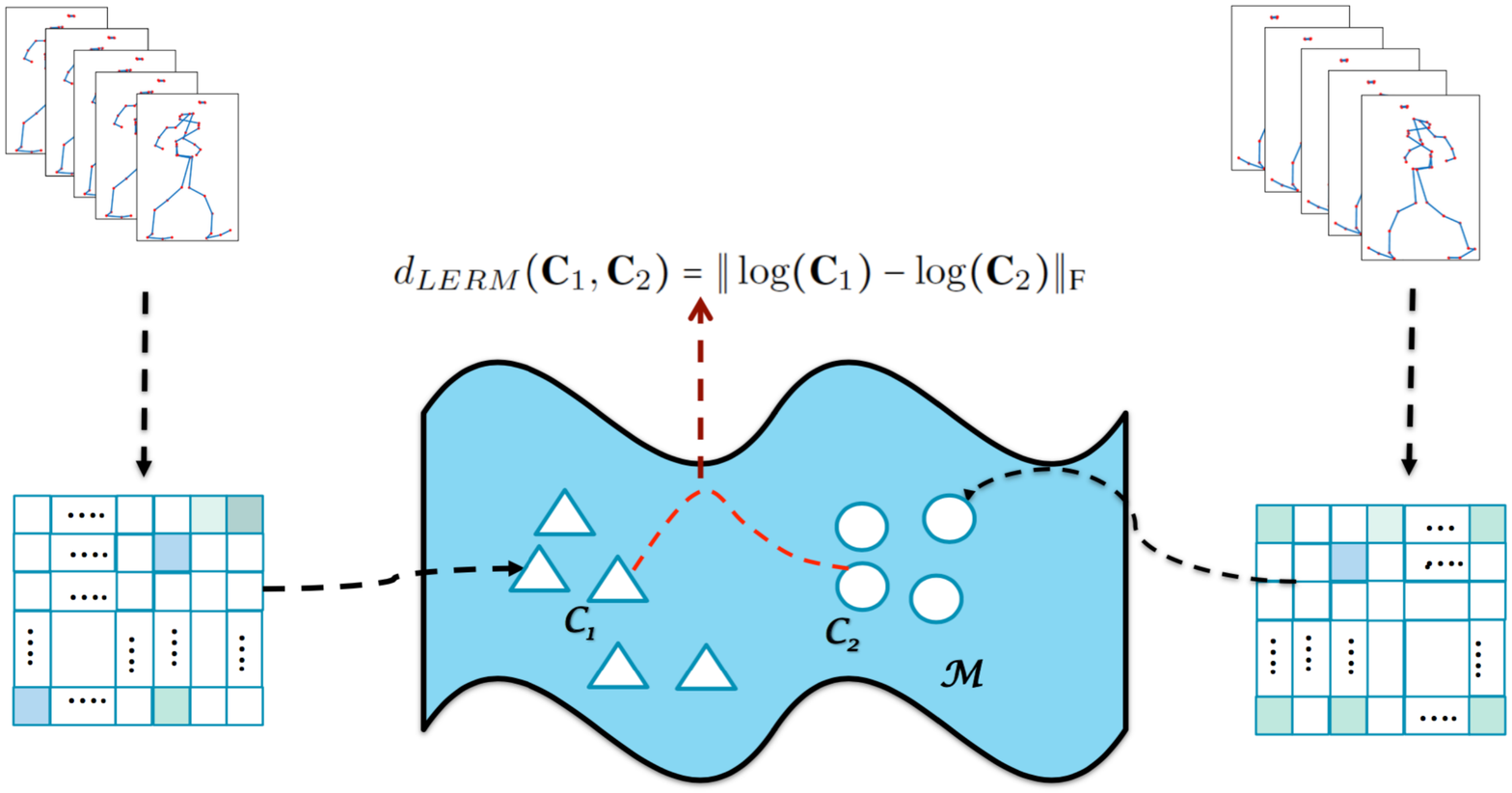}
\caption{\label{fig:manifold} Each body motion sequence is represented through a covariance matrix, which is a point on the Riemannian manifold of SPD matrices. Distance between sequences is then evaluated as the geodesic between points on the manifold}
\end{figure}

Figure~\ref{fig:manifold} shows the idea of capturing the body movement in a sequence through a covariance matrix which, in turn, is a point on the SPD manifold.

\section{Emotion Classification}\label{sect:classification}
The covariance matrix computed from skeleton data observed across a temporal window $[0,T]$ retains a signature of the emotional state of the observed person.
To perform emotion recognition, in the proposed approach, the covariance matrix computed from an unknown observation is compared with the Prototype Emotional Matrices (PEMs) representative of the target emotions (in the experiments reported in Sect.~\ref{sect:results}, five basic emotions are considered, namely, \textit{anger}, \textit{fear}, \textit{joy}, \textit{neutral}, \textit{sadness}). 
The unknown sequence is classified according to the emotion associated with the closest PEM. Computation of PEMs relies on extraction of representative examples from training data.
It should be noted that, according to what is described in Sect.~\ref{sect:spd}, the computation of the distance to PEMs as well as the identification of PEMs from training data should both take into account the fact that covariance matrices lie on the Riemannian manifold of SPD matrices $Sym^{+}_d$. This prevents the use of common tools adopted in Euclidean spaces to compute distances between points and cluster them.

%


Let $\left\lbrace \textrm{\textbf{C}}_i , \textrm{l}_i\right\rbrace_{i=1\ldots N}$ be a training set of labeled samples composed of covariance matrices $\textrm{\textbf{C}}_i$ and corresponding emotion labels $\textrm{l}_i \in \left\lbrace l_1,\ldots,l_E \right\rbrace$.
The emotion classification task acts like a function that associates with a generic element of $Sym^{+}_d$ its classification label $l \in \left\lbrace l_1,\ldots,l_E \right\rbrace$.
A possible solution would be to adopt a nearest-neighbor ($k$-NN) approach by comparing the covariance matrix to be classified to all the labeled covariance matrices in the training, and assigning to it the same label of the closest matrix (for instance, the LERM distance in~(\ref{eq:LERM}) can be used for the comparison).
A better solution, both in terms of computation and of generalization of training examples is to extract some representative prototypes from the training examples. Then, it would be possible to compare the covariance matrix to classify to these prototypes, instead of using all training examples.
Following this idea, we extract a PEM from each emotion class. This is achieved by computing, for each emotion class $l_i$ the Riemannian Center of Mass 
of all the training examples with label $l_i$.
Given a set of covariance matrices $\left\lbrace \textrm{\textbf{C}}_i \right\rbrace_{i=1\ldots N}$ on the Riemannian manifold $Sym^{+}_d$, the Riemannian Center of Mass, also referred to as \emph{Karcher mean} in the literature, is the point on $Sym^{+}_d$ that minimizes the sum of squared Riemannian distances:
\begin{equation}
\label{eq:karcher-mean}
\mathbf{\mu} = \arg\!\min_{\textrm{\textbf{C}} \in Sym^{+}_d} \sum_{i=1}^{N} d^2 \left(  \textrm{\textbf{C}}, \textrm{\textbf{C}}_i \right) \; ,
\end{equation}

\noindent being $d(\cdot)$ a suitable distance measure on the manifold.

It should be noted that, in case the LERM distance in~(\ref{eq:LERM}) is used, the Riemannian Center of Mass can be computed in closed form through the following expression~\cite{zhang:2016}:
\begin{equation}
\label{eq:karcher-mean-closed-form}
\mathbf{\mu} = \exp \left( \frac{1}{N} \sum_{i=1}^N \log \left( \textrm{\textbf{C}}_i  \right) \right) \; ,
\end{equation}

\noindent being $\exp(\cdot)$ and $\log(\cdot)$ the matrix exponential and logarithm operators, respectively. 
In this way, for the emotion corresponding to label $l_i$, the Prototype Emotional Matrix $\textrm{\textbf{Pem}}_{l_i}$ is computed as the Riemannian center of mass of all training samples $\left\lbrace \textrm{\textbf{C}}_k , \textrm{l}_k\right\rbrace$, such that $\textrm{l}_k = \textrm{l}_i$.
A generic covariance matrix to be classified is assigned the label corresponding to the closest $\textrm{\textbf{Pem}}_{l_i}$.
In doing so, the identification of Prototype Emotional Matrices as well as the classification of the emotion to be associated to a new covariance matrix rely on a measure of distance that preserves the inherent structure of the manifold. 


\section{Experiments}\label{sect:results}
\begin{figure*}[ht!]
\centering
\includegraphics[width=0.15\linewidth, viewport=7cm 9.2cm 14cm 20cm, clip]{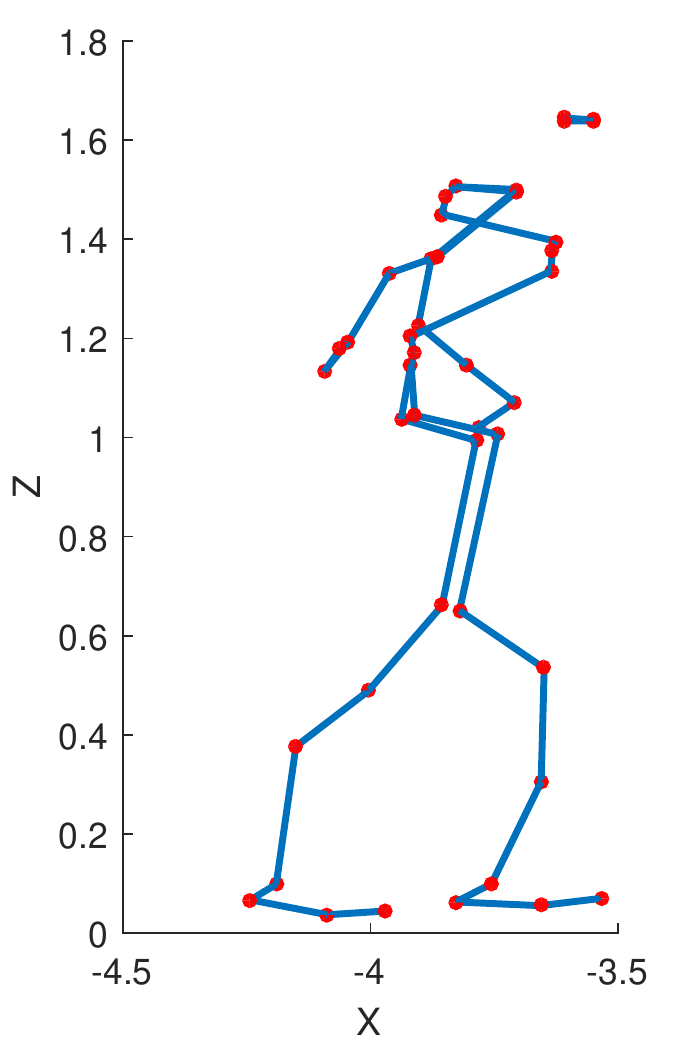}
\hfill
\includegraphics[width=0.2\linewidth, viewport=5.8cm 9.2cm 15cm 20cm, clip]{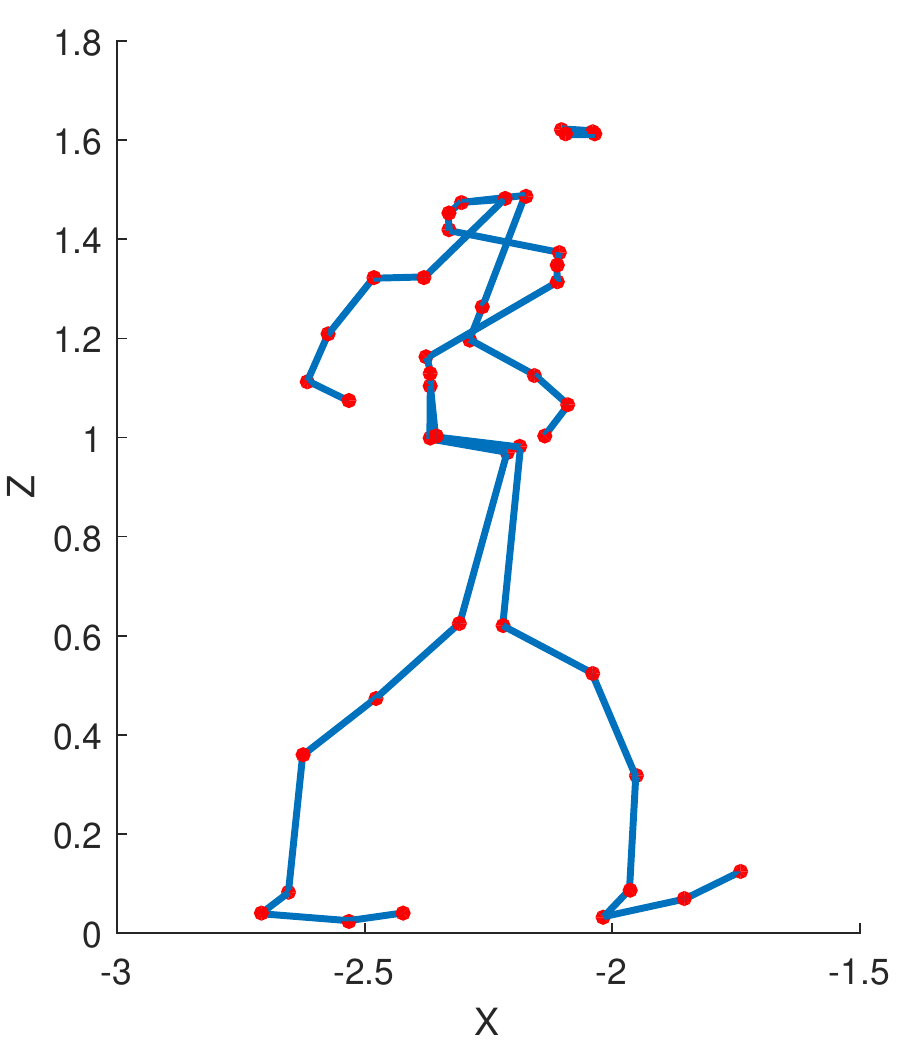}
\hfill
\includegraphics[width=0.15\linewidth, viewport=7cm 9.2cm 14cm 20cm, clip]{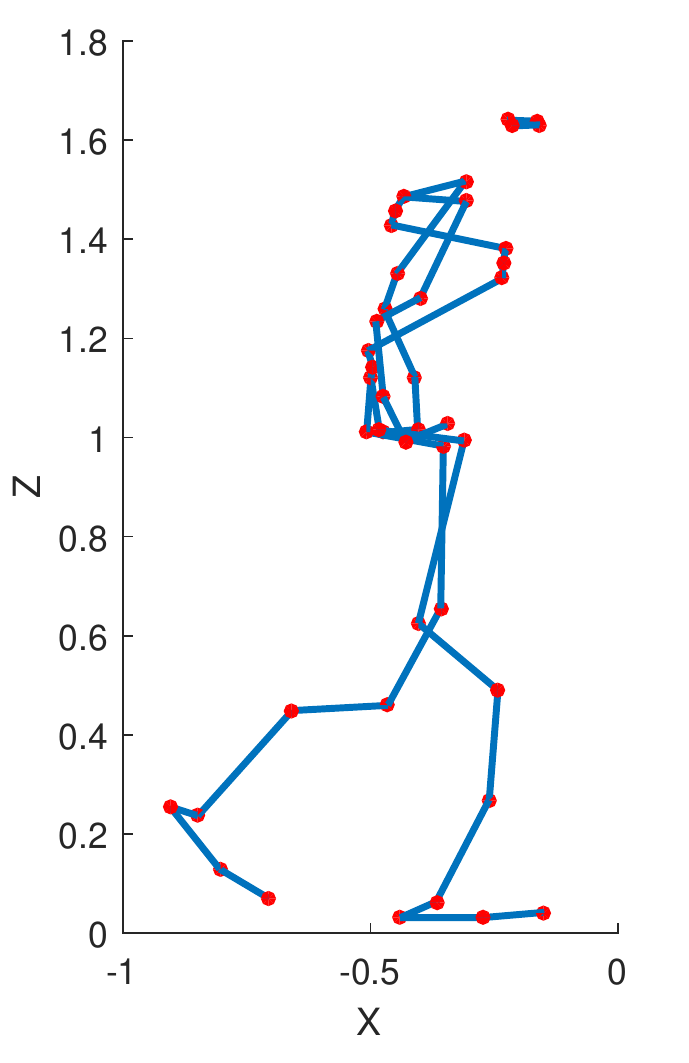}
\hfill
\includegraphics[width=0.15\linewidth, viewport=7cm 9.2cm 14cm 20cm, clip]{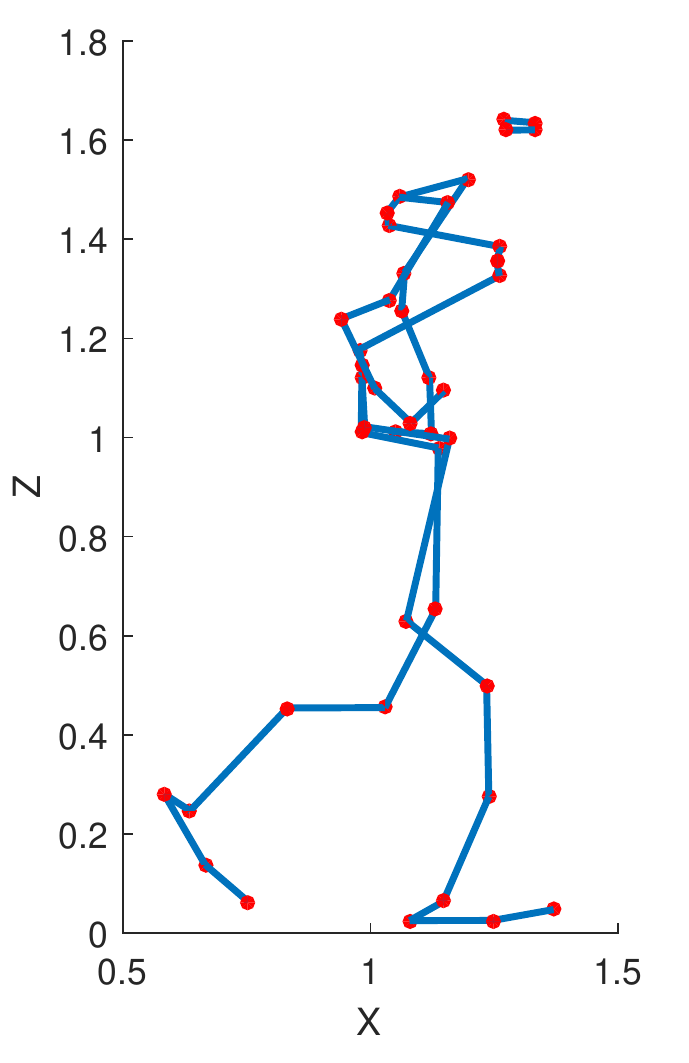}
\hfill
\includegraphics[width=0.15\linewidth, viewport=7cm 9.2cm 14cm 20cm, clip]{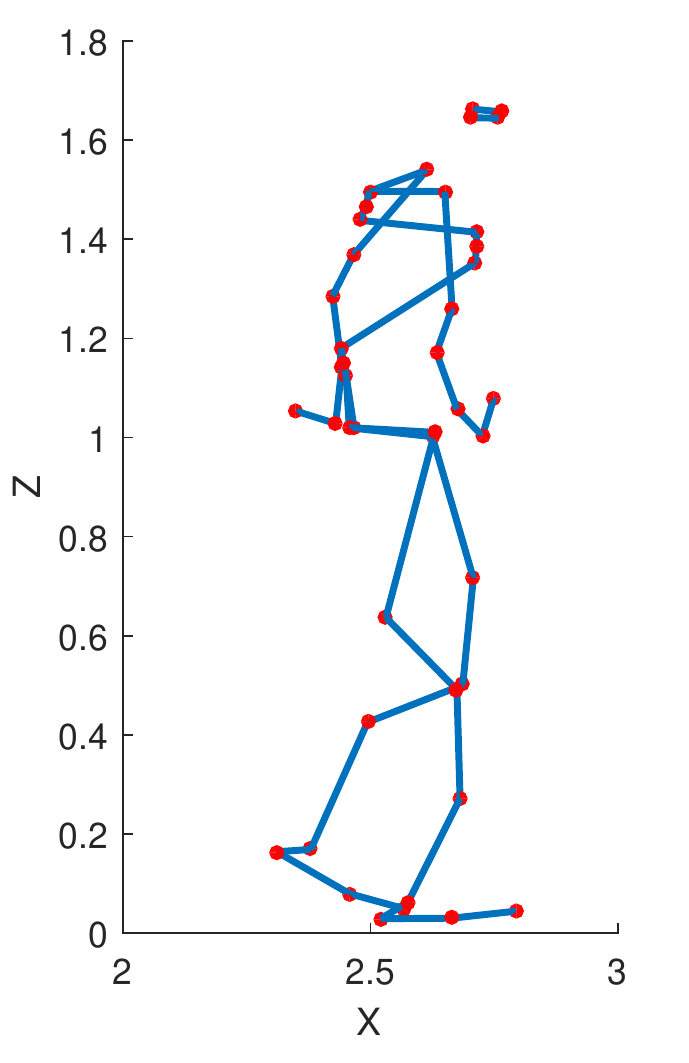}
\\
\includegraphics[width=0.19\linewidth, viewport=5.8cm 9.2cm 15cm 20cm, clip]{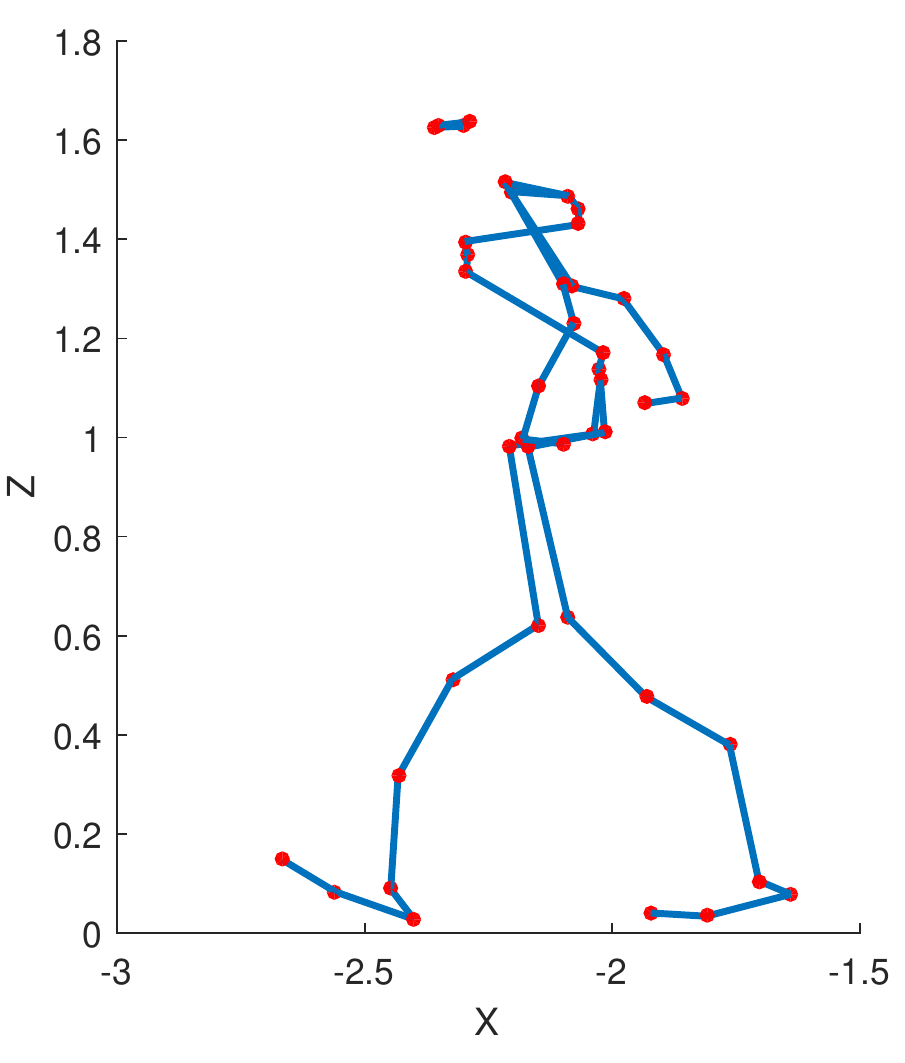}
\hfill
\includegraphics[width=0.19\linewidth, viewport=5.8cm 9.2cm 14.8cm 20cm, clip]{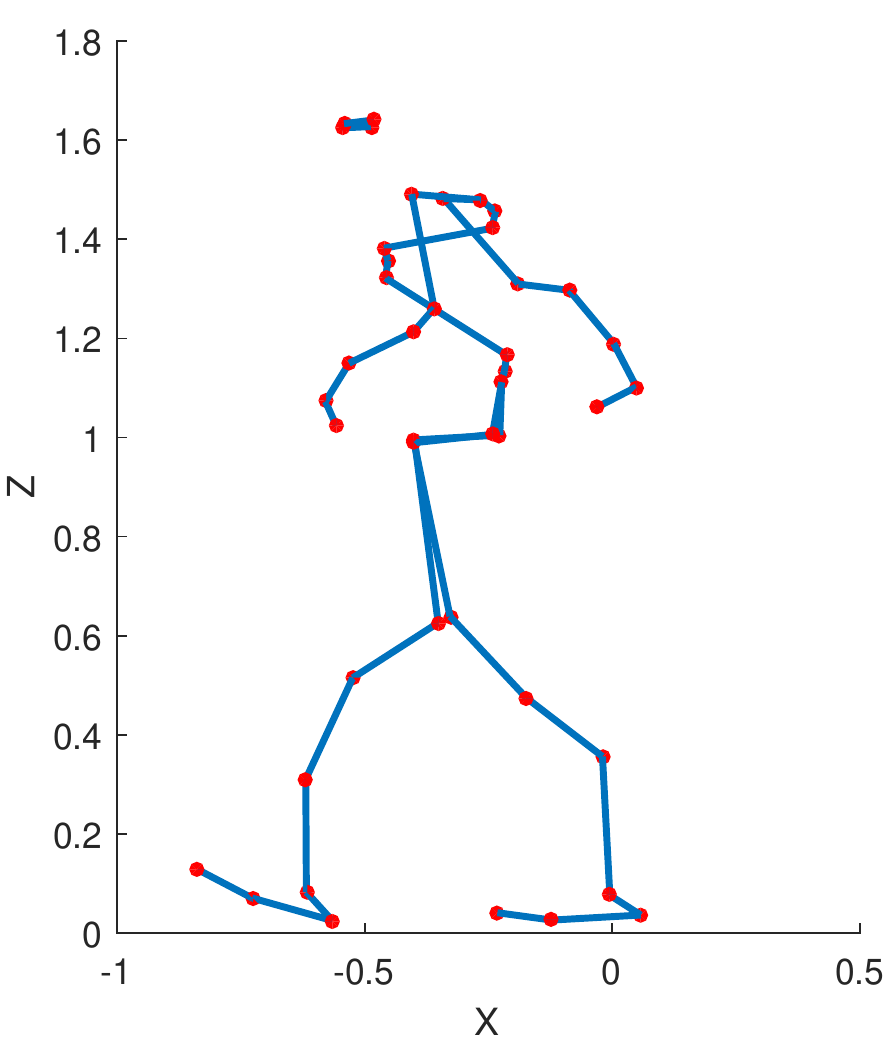}
\hfill
\includegraphics[width=0.15\linewidth, viewport=7cm 9.2cm 14cm 20cm, clip]{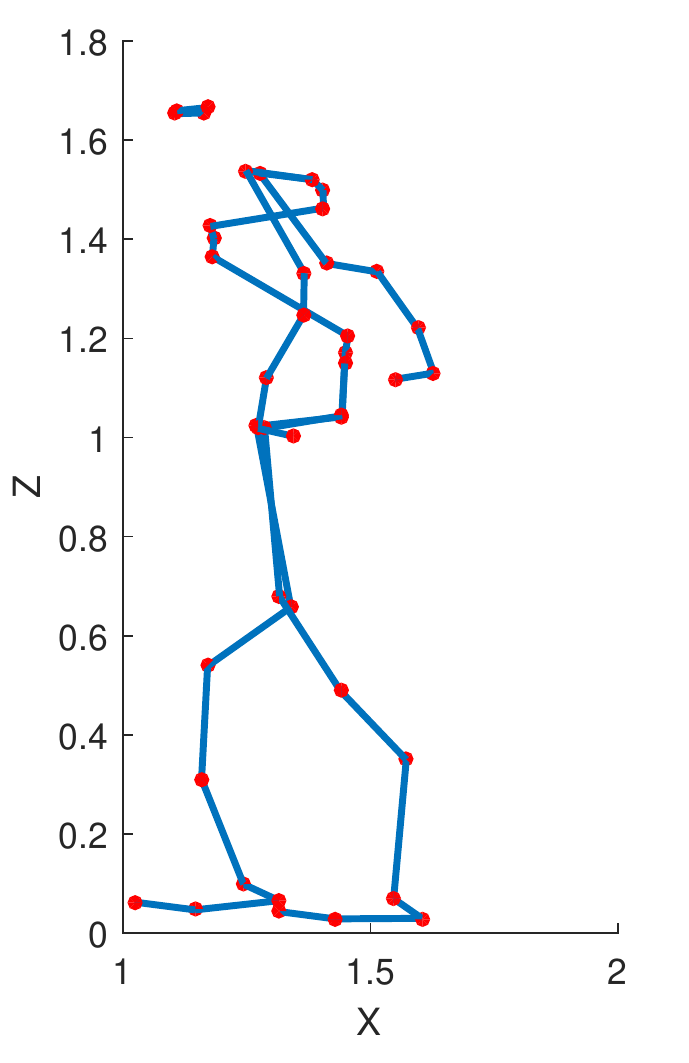}
\hfill
\includegraphics[width=0.19\linewidth, viewport=5.8cm 9.2cm 14.8cm 20cm, clip]{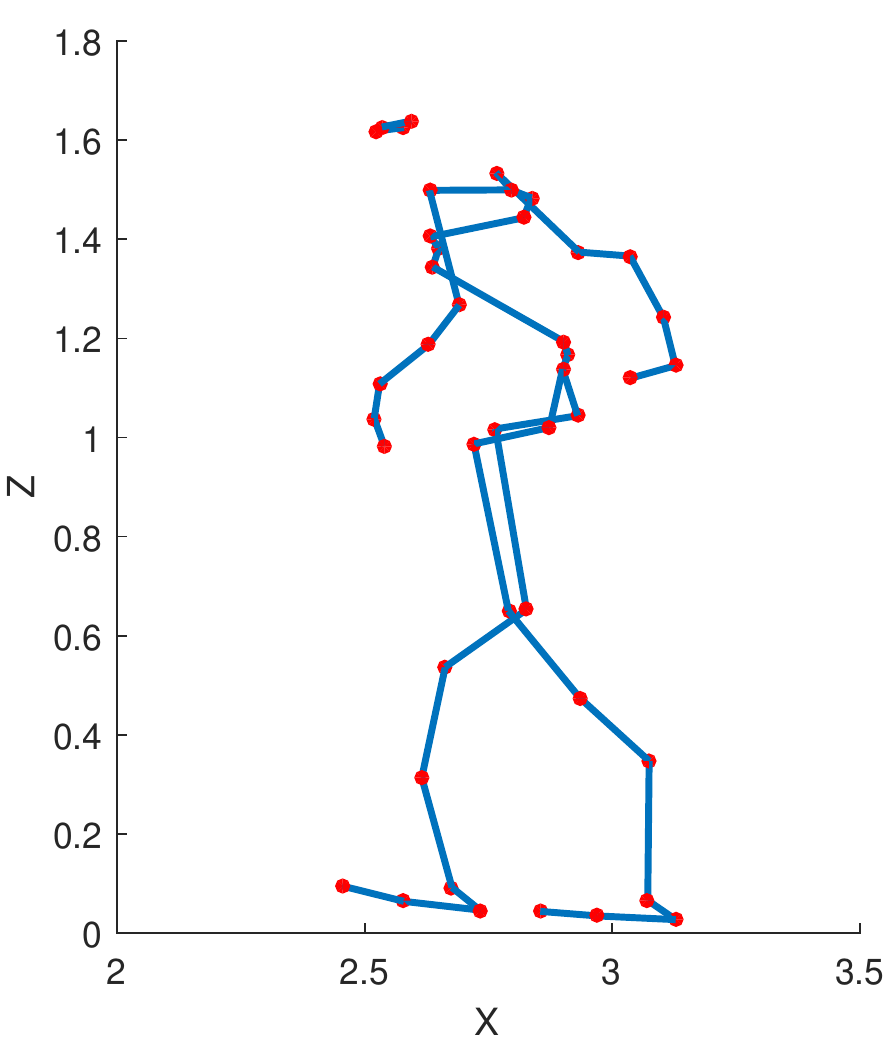}
\hfill
\includegraphics[width=0.15\linewidth, viewport=7cm 9.2cm 14cm 20cm, clip]{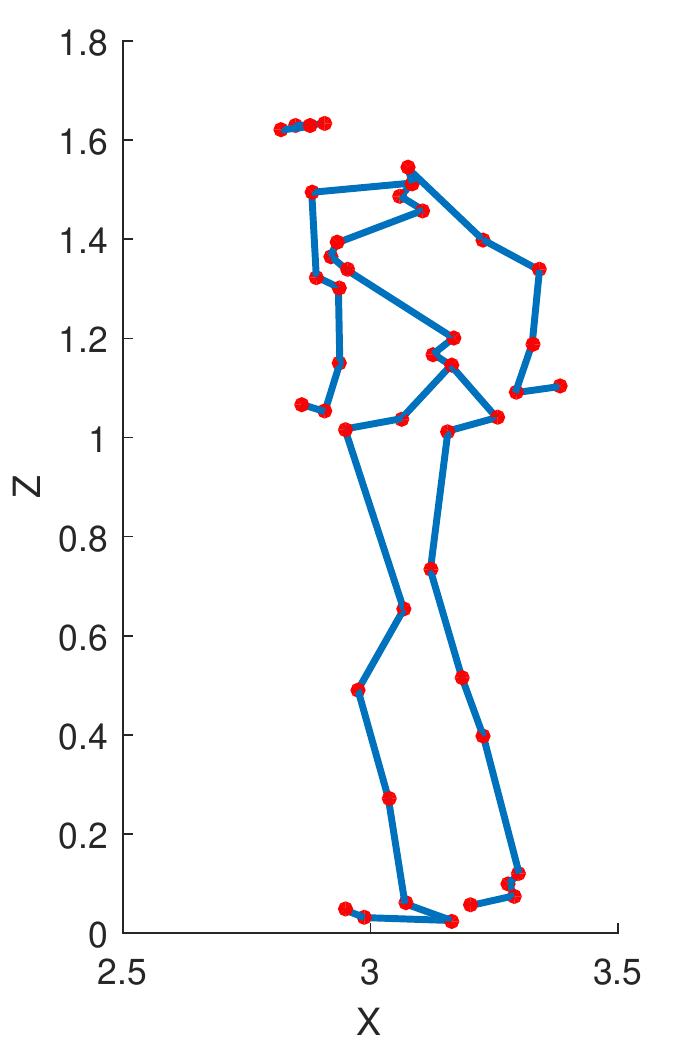}
\caption{\label{fig:mocap} Frames from a MoCap skeleton sequence of the P-BME dataset. In this example, an actor moves following a ``U'' shaped trajectory showing an \textit{anger} emotion. In the top row, the subject advances towards the turning point (plots from left-to-right); in the bottom row, the subject moves away from the turning point (plots from right-to-left). The changes in the moving direction at the turning point can be observed in the rightmost frame of both top and bottom rows. In each frame, the skeleton is represented by 43 joints. Connections between joints are shown (except for the four joints of the head) to evidence the silhouette of the body and the limbs}
\end{figure*}

Experiments have been performed on the Body Motion-Emotion dataset (P-BME), 
that has been acquired at the Cognitive Neuroscience Laboratory (INSERM U960 - Ecole Normale Sup\'{e}rieure) in Paris~\cite{hicheur:2010}. 
It includes Motion Capture (MoCap) 3D data sequences 
recorded at a high frame rate (120 frames per second) by an Opto-electronic Vicon V8 MoCap system wired to 24 cameras. 
The body movement is captured by using 43 landmarks that are positioned at joints and other parts of the body as illustrated in Fig.~\ref{fig:mocap}. 
To create the dataset, 8 subjects (professional actors) were instructed to walk following a predefined ``U'' shaped path that includes forward-walking, turn, and coming back (Fig.~\ref{fig:mocap}).
For each acquisition, actors move along the path performing one emotion out of a set of five different emotions, namely, \textit{anger}, \textit{fear}, \textit{joy}, \textit{neutral}, and \textit{sadness}. So, each sequence is associated with one emotion label. In doing so, the emotional gait patterns show to be characterized by different walking velocity, wrist velocity and acceleration, body and head postures. Each actor performed at maximum five repetitions of a same emotional sequence for a total of 156 instances.
Though there is some variation from subject to subject, the number of examples is well distributed across the different emotions: 29 \textit{anger}, 31 \textit{fear}, 33 \textit{joy}, 28 \textit{neutral}, 35 \textit{sadness}.

\subsection{Results and Comparative Evaluation}
Experiments on the P-BME dataset were performed by using a \textit{leave-one-subject-out} cross validation protocol. With this solution, iteratively, all the emotion sequences of a subject are used for test, while all the sequences of the remaining subjects are used for training. As discussed in Sect.~\ref{sect:classification}, the training sequences are used to perform supervised clustering in the five emotional classes. This is obtained by first computing the Riemannian center of mass of each emotion class and retaining it as representative element of the class. Then, nearest-neighbor classification of the test sequence is performed by computing the LERM distance to these representative elements. A confusion matrix is thus computed for each fold. Averaging such matrices across the eight folds (also weighting each matrix according to the relative number of test examples, which is different from subject to subject) we obtain the overall results reported in Table~\ref{tab:cm-karcher}. It can be observed the diagonal dominance of the matrix (average positive classification of about 71\%), with the best results scored by \textit{neutral} and \textit{anger} (about 80\%), followed by \textit{sadness} and \textit{fear} (about 68\%), with the lowest accuracy for \textit{joy} (about 58\%).

\begin{table}[!ht]
\caption{\label{tab:cm-karcher} P-BME dataset: Emotion recognition accuracy obtained using the Riemannian center of mass (results in percentage). Average accuracy is 71.12\%}
\centering
\renewcommand{\arraystretch}{1.1}
\footnotesize
\begin{tabular}{l||c|c|c|c|c}
\hline
   		& \quad \textbf{Anger} \quad & \quad \textbf{Fear} \quad & \quad \textbf{Joy} \quad & \quad \textbf{Neutral} \quad & \quad \textbf{Sadness} \quad \\
\hline
\hline
\textbf{Anger} \quad & \textbf{79.31} & 3.45 & 13.79 & 0.00 & 3.45 \\
\hline
\textbf{Fear}  \quad & 3.57 & \textbf{67.86} & 10.71 & 0.00 & 17.86 \\
\hline			
\textbf{Joy} \quad   & 3.23 & 6.45 & \textbf{58.06} & 9.68 & 22.58 \\
\hline
\textbf{Neutral} \quad & 6.06 & 0.00 & 0.00 & \textbf{81.82} & 12.12 \\
\hline
\textbf{Sadness} \quad & 2.86 & 20.00 & 2.86 & 5.71 & \textbf{68.57} \\
\hline
\end{tabular}
\end{table}

We also performed experiments by using nearest-neighbor (NN) classification with respect to all the training sequences, without reducing them with any clustering operation. 
In addition to be much more computational demanding, this classification scores substantially lower results as reported in Table~\ref{tab:cm-nn} (the average of the diagonal values decreases to about 51\%). This confirms us the intuition that performing the Riemannian center of mass on the training sequences can reduce the effects induced by outliers included in the training examples that were provided for each emotion. 

To also validate the importance of measuring distances between covariance matrices using geodesic distances on the manifold, compared to standard matrix norm computation, we performed NN-classification using the Frobenius norm of the difference between covariance matrices. This resulted in an average classification of 43.4\% which is more than 7\% less than the result obtained using LERM in Table~\ref{tab:cm-nn}.   

\begin{table}[!ht]
\caption{\label{tab:cm-nn} P-BME dataset: Emotion recognition accuracy obtained using a nearest-neighbor approach (results in percentage). Average accuracy is 50.74\%}
\centering
\renewcommand{\arraystretch}{1.1}
\begin{tabular}{l||c|c|c|c|c}
\hline
   		& \quad \textbf{Anger} \quad & \quad \textbf{Fear} \quad & \quad \textbf{Joy} \quad & \quad \textbf{Neutral} \quad & \quad \textbf{Sadness} \quad \\
\hline
\hline
\textbf{Anger} \quad   & \textbf{41.38} & 0.00 & 3.45 & 31.03 & 24.14 \\
\hline
\textbf{Fear} \quad    & 0.00 & \textbf{67.86} & 7.14 & 3.57 & 21.43 \\
\hline			
\textbf{Joy} \quad     & 0.00 & 3.23 & \textbf{16.13} & 32.26 & 48.39 \\
\hline
\textbf{Neutral} \quad & 0.00 & 0.00 & 0.00 & \textbf{45.45} & 54.55 \\
\hline
\textbf{Sadness} \quad & 2.86 & 11.43 & 0.00 & 2.86 & \textbf{82.86} \\
\hline
\end{tabular}
\end{table}

Finally, we performed a user based test in order to evaluate the performance of the proposed classification method in comparison with a human-based judgment. In this test, thirty-two naive individuals (with heterogeneous age and no experience in human emotion classification) were asked to perform a force-choice task. Participants were seated in front of a computer screen, 
and videos were presented following a semi-randomized block design, with nature of emotion randomly presented for each actor. The order of the presentations of the video clips for each actor was also counter-balanced. Participants were required to categorize the observed motion sequences in one of the five emotional categories within 5$secs$ after the end of the video presentation, using the Geneva Emotional Wheel (GEW)~\cite{scherer:2005}. 
The task was a force choice situation in which the participants had to choose between one of five emotions: \textit{anger}, \textit{fear}, \textit{joy}, \textit{sadness} or \textit{neutral}. Table~\ref{tab:hum-classification} reports the scores obtained for emotion classification based on RGB videos by the human evaluators. The results reveal an average value of about 74\%, which is just 3\% over the average result found in Table~\ref{tab:cm-karcher}.
It is relevant to note that the user based test being based on RGB videos provides to the users much more information for evaluation, including the actor's face. Notably, our method is capable to score comparable results based on the skeleton joints only.

\begin{table}[!ht]
\caption{\label{tab:hum-classification} P-BME dataset: Emotion recognition of body motion by human evaluator}
\centering
\renewcommand{\arraystretch}{1.1}
\begin{tabular}{c|c|c|c|c||c}
\hline
\textbf{Anger} \quad & \quad \textbf{Fear} \quad & \quad \textbf{Joy} \quad & \quad \textbf{Neutral} \quad & \quad \textbf{Sadness} \quad & \quad \textbf{Average} \\
\hline
\hline
84.0 & 81.5 & 73.5 & 65.0 & 67.0 & 74.2 \\
\hline
\end{tabular}
\end{table}

\section{Conclusions}\label{sect:conclusions}
In this work, we focus on 3D dynamic sequences of the body skeleton and propose a new method to relate automatically human body movements to inner sensorial emotion. This is obtained by first representing the 3D evolution of the skeleton joints across time by using a covariance matrix. Then, we account for the fact that these matrices lay in the non-linear Riemannian manifold of SPD matrices. Exploiting geodesic distances and geometric average computation on the manifold, emotion classification is performed.
Results obtained in the experiments show an average recognition of about 71\% for the proposed method, which is comparable with the average score produced by human evaluation.  Notably, our results have been obtained using only joints information, while humans evaluators exploited the richer RGB video channel.
The covariance matrix captures the dependence of locations of different joints on one another during the performance of an human action. The covariance matrix does not capture the order of motion in time. 
Future work will address more advanced approaches for modeling the temporal evolution and machine learning and classification methods on a non-linear manifold. We will also investigate the generalization of the method by applying it to other types of voluntary motor actions besides walking (e.g., cycling, running, or cooking a meal).

\section{Acknowledgements}
This work has been partially supported by PIA, ANR (grant ANR-11-EQPX-0023), European Founds for the Regional Development (Grant FEDER-Presage 41779). We thank Julie Grèzes and Alain Berthoz (INSERM U960, France) and Halim Hicheur (University of Fribourg, Switzerland) for the possibility to use their data set. During the preparation of this paper (July 2016), M. Daoudi enjoyed excellent working conditions at the University of Florence, Italy.


\end{document}